\documentclass[11pt,a4paper]{article}
\usepackage[hyperref]{acl2021}
\usepackage{times}
\usepackage{latexsym}

\usepackage{amsmath}
\usepackage{amssymb}
\usepackage{graphicx}
\usepackage{float}
\DeclareMathOperator*{\argmax}{argmax}

\usepackage{microtype}

\aclfinalcopy 
\def\aclpaperid{1646} 

\title{Word Embedding Transformation for Robust Unsupervised Bilingual Lexicon Induction}

\author{
Hailong Cao \rm{and} \textbf{Tiejun Zhao}
\\
Harbin Institute of Technology \\
caohailong@hit.edu.cn,
tjzhao@hit.edu.cn
}

\date{}

\begin{document}
\maketitle
\begin{abstract}
Great progress has been made in unsupervised bilingual lexicon induction (UBLI) by aligning the source and target word embeddings independently trained on monolingual corpora. The common assumption of most UBLI models is that the embedding spaces of two languages are approximately isomorphic. Therefore the performance is bound by the degree of isomorphism, especially on etymologically and typologically distant languages. To address this problem, we propose a transformation-based method to increase the isomorphism. Embeddings of two languages are made to match with each other by rotating and scaling. The method does not require any form of supervision and can be applied to any language pair. On a benchmark data set of bilingual lexicon induction, our approach can achieve competitive or superior performance compared to state-of-the-art methods, with particularly strong results being found on distant languages.
\end{abstract}

\section{Introduction}

Unsupervised bilingual lexicon induction (UBLI) is the task of inducing word translations from only monolingual corpora of two languages.
UBLI is both theoretically interesting and practically important for many low-resource scenarios where cross-lingual training data is not available
 \citep{DBLP:conf/iclr/ArtetxeLAC18,lample-etal-2018-phrase,ruder-etal-2019-unsupervised}.
Recent work shows that good performance can be achieved by aligning the source and target word embeddings independently trained on monolingual corpora \citep{zhang-etal-2017-adversarial,conneau2017word,artetxe-etal-2018-robust,alvarez-melis-jaakkola-2018-gromov,hoshen-wolf-2018-non,xu-etal-2018-unsupervised}.
In particular, the source embeddings space is mapped into the target embedding space so that words and their translations are close to each other in the shared embedding space.

The common assumption of most UBLI models is that the embedding spaces of source and target languages are approximately isomorphic. Therefore the performance is bound by the degree of isomorphism. Near-zero UBLI results have been shown for pairs of etymologically and typologically distant languages \citep{sogaard-etal-2018-limitations,vulic-etal-2019-really,patra-etal-2019-bilingual,glavas-vulic-2020-non}.
To address this problem, we propose a method for robust UBLI based on embedding transformation. The method does not require any form of supervision and can be applied to any language pair. In addition, the method is extremely easy to implement. Our contributions include:
\begin{itemize}

  \item We propose a novel method to make the word embeddings of two languages to match with each other. This is realized by rotating and scaling the embeddings.
  \item We achieved competitive or superior performance compared to state-of-the-art methods. Our code will be made public soon.
 
\end{itemize}
\section{Background}

As a background and a baseline, in this section, we briefly describe the self-learning model VecMap\footnote{https://github.com/artetxem/vecmap}\citep{artetxe-etal-2018-robust}.
Let $X\in R^{m\times d}$ and $Z\in R^{n\times d}$ be the word embedding matrices in two languages, so that their $i$th row $X_{i*}$ and $Z_{i*}$ denote the $d$-dimensional embeddings of the $i$th word in their respective vocabularies. We assume that both the vocabulary size $m$ and $n$ are bigger than $d$.
The goal of the VecMap is to learn the linear transformation matrices $W_X\in R^{d\times d}$ and $W_Z\in R^{d\times d}$ so the mapped embeddings $XW_X$ and $ZW_Z$ are in the same cross-lingual space.

The method consists of four sequential steps which are:
\begin{itemize}
\item[1.] Embedding normalization.
\item[2.] Initialization.
\item[3.] Self-learning.
\item[4.] Refinement.
\end{itemize}

Firstly, VecMap length normalizes the embeddings, then mean centers each dimension, and then length normalizes them again.
Secondly, the initialization builds a dictionary $D$ based on similarity matrices $XX^{T}$ and $ZZ^{T}$. $D$ is a $m$-by-$n$ sparse matrix where $D_{ij} = 1$ if the $j$th word in the target language is a translation of the $i$th word in the source language.
Thirdly, with the initial dictionary $D$, the self-learning iterates the following two steps:
\begin{itemize}
\item Compute the optimal orthogonal mapping maximizing the similarities for the current dictionary $D$:
\begin{equation}\label{selflearning}
\mathop{\argmax}_{W_X,W_Z}{\sum_i\sum_j{D_{ij}((X_{i*}W_X) \cdot (Z_{i*}W_Z))}}
\end{equation}
An optimal solution is given by $W_X=U$ and $W_Z=V$, where $USV^T = X^TDZ$ is the singular value decomposition of $X^TDZ$.
\item Induce a new dictionary $D$ based on the similarity matrix of the mapped embeddings $XW_XW_Z^TZ^T$.
\end{itemize}

Finally, the refinement step further improves the resulting mapping through symmetric re-weighting \citep{artetxe2018aaai}.
VecMap is one the most strong and robust UBLI models. In addition, it is faster than GAN-based  methods. Please refer to the original paper \citep{artetxe-etal-2018-robust} for more detail  of VecMap.

\begin{table*}[htbp]
\small
\centering
\begin{tabular}{lllllllllll}
\hline
 & En-Ru & Ru-En & En-Zh & Zh-En & En-Ja & Ja-En & En-Vi & Vi-En & En-Th & Th-En\\
\hline
\citet{zhou-etal-2019-density} & 47.3 & 63.5& \textbf{41.9} & 37.7 & 50.7 & \textbf{35.2} & - & -& - & -\\
\citet{li-etal-2020-simple} & - & -& 37.33 & 35.27 & 48.87 & 33.08& \textbf{47.6} & \textbf{55.53}& 21.60 & 13.64\\
MUSE & 44.33 & 60.33& 32.53 & 30.26 & 0.00 & 0.00 & 0.13 & 0.00& 0.00 & 0.00\\
VecMap & 48.73 & \textbf{65.47}& 0.00 & 0.00 & 47.09 & 32.94& 0.13 & 0.20& 0.00 & 0.00\\
Proposed method & \textbf{49.00} & 64.33 &39.47 & \textbf{40.60} & \textbf{51.75} & 33.77 & 46.40 & 53.73 & \textbf{23.67} & \textbf{18.64}\\
\hline
\end{tabular}
\caption{\label{distant}
The accuracy of UBLI on distant language pairs. Rs means the random seed. En is English, Ru is Russian, Zh is traditional Chinese, Ja is Japanese, Vi is Vietnamese and Th is Thai. The bold numbers indicate the best results of all methods.
}
\end{table*}
\section{Proposed Method}

In Eq.\ref{selflearning}, the orthogonal transformations $W_X$ and $W_Z$ do not affect the topology of embeddings spaces, so the performance is bound by the degree of isomorphism. In order to increase isomorphism for better UBLI, we propose to preprocess the embeddings based on a transformation which is applied between the normalization step and the initialization step of VecMap. 
In particular, the transformation is composed of a rotation and a scaling.

\subsection{Rotating}
As noted by \citet{artetxe-etal-2018-robust}, the underlying difficulty of increasing the isomorphism without any supervision is that the word embedding matrices $X$ and $Z$ are unaligned across both axes. 

In order to overcome this challenge, we propose a fully unsupervised approach to rotate the original embeddings $X$ and $Z$ into a same cross-lingual space where the isomorphism can be increased. The intuition of our method is as follows.

Let us assume that the embedding spaces are perfectly isometric so that $X$ and $Z$ are equivalent up to a permutation of their rows and a rotation. Mathematically, we formalize the perfect isometry as $m=n$ and:
\begin{equation}
X=PZO
\end{equation}
where $P\in R^{n\times n}$ is an unknown row permutation matrix such that every row and column contains precisely a single 1 with 0s everywhere else. $O\in R^{d\times d}$ is an unknown orthogonal rotation matrix. As both $P$ and $O$ are orthogonal, it can be proved that components of the singular value decomposition of $X$ and $Z$ have the following relations:
\begin{equation}\label{svd}
\begin{aligned}
U_X&=PU_Z\\
S_X&=S_Z\\
V_X^T&=V_Z^TO
\end{aligned}
\end{equation}
where $U_XS_XV_X^T=X$ and $U_ZS_ZV_Z^T=Z$ are the singular value decomposition of $X$ and $Z$ respectively.
It follows immediately that:
\begin{equation}
U_XS_X=PU_ZS_Z
\end{equation}
Since $XV_X=U_XS_X$ and $ZV_Z=U_ZS_Z$, we can have:
\begin{equation}\label{rotated}
XV_X=PZV_Z
\end{equation}
This indicates that rotated embedding matrices $XV_X$ and $ZV_Z$ are equivalent up to a row permutation and therefore are aligned across their dimensions. In other words, even though both the permutation matrix $P$ and the rotation matrix $O$ are unknown, the row vectors of $XV_X$ and $ZV_Z$ are in the same cross-lingual $d$-dimensional space.

In practice, the isometry requirement will not hold exactly, but it can be assumed to hold approximately, as the very same problem of mapping two embedding spaces without supervision would otherwise be hopeless. So we can hypothesize that, to some extend, $XV_X$ and $ZV_Z$ are aligned across their $j$th dimension($1\leq j\leq d$) on practical data. This hypothesis is verified by our experiments(section \ref{rotation}). While the dimension alignment is far from being useful on its own, it can enable us to increase the isomorphism by matching the norms of each dimension.
\subsection{Scaling}

If $X$ and $Z$ are perfectly isometric, Eq.\eqref{rotated} shows that the $j$th column $XV_X$ and $ZV_Z$ are equivalent up to a row permutation. Since row permutation only changes the orders of elements in a column, the $L$-$2$ norms of the $j$th column of $XV_X$ and $ZV_Z$ should be equal. For example, the norms of column vectors $(a,b,c)^T$ and $(b,a,c)^T$ are both $\sqrt{a^2+b^2+c^2}$. 
In practice, experiments show that the norms of each column are approximately equal on similar languages such as English and French(section \ref{norms}). However, the norms are far from equal on distant language pairs. In order to increase the isomorphism, we propose to make them match with each other.

Since $XV_X=U_XS_X$ and $ZV_Z=U_ZS_Z$, the $L$-$2$ norms of the $j$th column of $XV_X$ and $ZV_Z$ are ${(S_X)}_{jj}$ and ${(S_Z)}_{jj}$ respectively.
To bridge the gap between them, we jointly scale each column of the rotated embedding matrices as follows:
\begin{equation}\label{gm}
\begin{aligned}
X^\prime&=XV_X\frac{\sqrt{S_Z}}{\sqrt{S_X}}\\
Z^\prime&=ZV_Z\frac{\sqrt{S_X}}{\sqrt{S_Z}}
\end{aligned}
\end{equation}
where the functions of square root and division operate element-wise across the diagonal of matrix $S_X$ and $S_Z$. In this way, the norms of $j$th dimension of both $X^\prime$ and $Z^\prime$ are $(\sqrt{S_XS_Z})_{jj}$ which is the geometric mean of ${(S_X)}_{jj}$ and ${(S_Z)}_{jj}$.
\subsection{Vocabulary cutoff}
However, simply replacing $X$ and $Z$ by $X^\prime$ and $Z^\prime$ only worked for a few language pairs in our preliminary experiments. More robust performance can be achieved if we limit the SVD procedure to the top 4,000 frequent words for both languages. The reason might be less frequent words can be expected to be noisier and the isomorphism degree of frequent words should be higher than that of less frequent words. Specifically, instead of applying the transformation shown in Eq.\eqref{gm}, we transform the embeddings as follows:
 
\begin{equation}\label{gm4k}
\begin{aligned}
X^{\prime\prime}&=X\hat{V}_X\frac{\sqrt{\hat{S}_Z}}{\sqrt{\hat{S}_X}}\\
Z^{\prime\prime}&=Z\hat{V}_Z\frac{\sqrt{\hat{S}_X}}{\sqrt{\hat{S}_Z}}
\end{aligned}
\end{equation}
where $\hat{U}_X\hat{S}_X\hat{V}_X^T=X[:4000]$ and $\hat{U}_Z\hat{S}_Z\hat{V}_Z^T=Z[:4000]$ are singular value decompositions of top 4,000 rows of $X$ and $Z$ respectively.

\subsection{Re-normalization} 
In VecMap, embeddings are pre-processed by first applying length normalization then mean center each dimension, and then length normalize again to ensure that the final embeddings have a unit length \citep {xing-etal-2015-normalized}. When normalized embedding matrices $X$ and $Z$ are transformed into $X^{\prime\prime}$ and $Z^{\prime\prime}$ by Eq.\eqref{gm4k}, the length of embeddings may not be one any more. So we normalize $X^{\prime\prime}$ and $Z^{\prime\prime}$ again by the default normalization procedure of VecMap. Then $X$ and $Z$ are discarded and the rest steps of VecMap are performed on $X^{\prime\prime}$ and $Z^{\prime\prime}$.
\section{Experiments}
\subsection{Setup}

We report results on the widely used MUSE dataset\footnote{https://github.com/facebookresearch/MUSE}, which consists of 300-dimensional FastText monolingual embeddings pretrained on Wikipedia \citep{bojanowski-etal-2017-enriching}, and dictionaries for many language pairs divided into train and test sets. All vocabularies are trimmed to the 200K most frequent words. The CSLS metric is used to induce the best translation for each source word. The accuracy of the induced lexicon is based on comparison with a gold standard. We run each experiment 3 times but with different random seeds, then pick the one with the highest average cosine similarity between these deemed translations as the final result. This unsupervised model selection(Ums) criterion has shown to correlate well with UBDI performance \citep{conneau2017word}. Our baselines are VecMap and MUSE. In all of our experiments, all hyper-parameters are set to the default values.

For comparison, we also include the results reported by two recent papers which achieved particularly strong performance for distant languages \citep{zhou-etal-2019-density,li-etal-2020-simple}.
\subsection{Results}
Table~\ref{distant} shows the UBLI results for distance language pairs.
Our method obtains better results than the VecMap on almost all language pairs with particularly strong performance being observed on En-Zh and En-Th. For example, our method has accuracies of 23.67\% in En-Th and 18.64\% in Th-En, both of which are very encouraging.

The performance of our method on En-Ja is comparable with that of \citet{zhou-etal-2019-density}. Given that their method utilized the identical strings in both languages, the results we achieved by fully unsupervised method are very promising.

Table~\ref{similar} shows the results for similar language pairs. Clearly, our method did not scarify the performance for similar language pairs.

\begin{table}
\small
\centering
\resizebox{\columnwidth}{!}{
\begin{tabular}{lllllll}	
\hline
 & En-Es & Es-En & En-De & De-En& En-Fr & Fr-En\\
\hline
MUSE &82.06& 84.26 & 74.80 & 72.73& 82.13 & 81.53\\
VecMap&82.20& 84.47 & 75.00 & 74.33& 82.40 & 83.67\\
Proposed&82.40& 84.60 & 75.40 & 74.20 & 82.27 & 83.47\\
\hline
\end{tabular}}
\caption{\label{similar}
The accuracy of UBLI on similar language pairs. Es is Spanish, De is German and Fr is French.
}
\end{table}
\subsection{The effect of rotation}\label{rotation}

As shown in Eq.\eqref{gm4k}, our transformation can be described as a rotation and a scaling. The goal of the rotation is to transform the original embeddings into a common space. To test the effect of rotation, we calculate the average cosine similarity across the gold standard dictionary on the original embeddings $X$ and $Z$, and the rotated embeddings $X\hat{V}_X$ and $Z\hat{V}_Z$ respectively.
Intuitively, the similarity of a word and its translation should be higher in the common space than that in the original space.
Table~\ref{dicsim4} shows the results. 
On both similar and distant language pairs, we observe much higher similarities on the rotated embeddings. 
So the rotation procedure does capture some cross-lingual signal based on which the isomorphism can be increased.

\begin{table}
\small
\centering
\begin{tabular}{lllll}
\hline
 & En-Es & En-De & En-Zh & En-Ja\\
\hline
Original & 0.0155 & 0.0195 &0.0128 & 0.0141\\
Rotated & 0.0893 & 0.0638 & 0.0775& 0.0622\\
\hline
\end{tabular}
\caption{\label{dicsim4}
Average cosine similarity on original embeddings and rotated embeddings.
}
\end{table}
\subsection{The singular values}\label{norms}
\begin{figure}[H]
\vspace{-0.5cm}
\centering
\includegraphics[height=5cm, width=7cm]{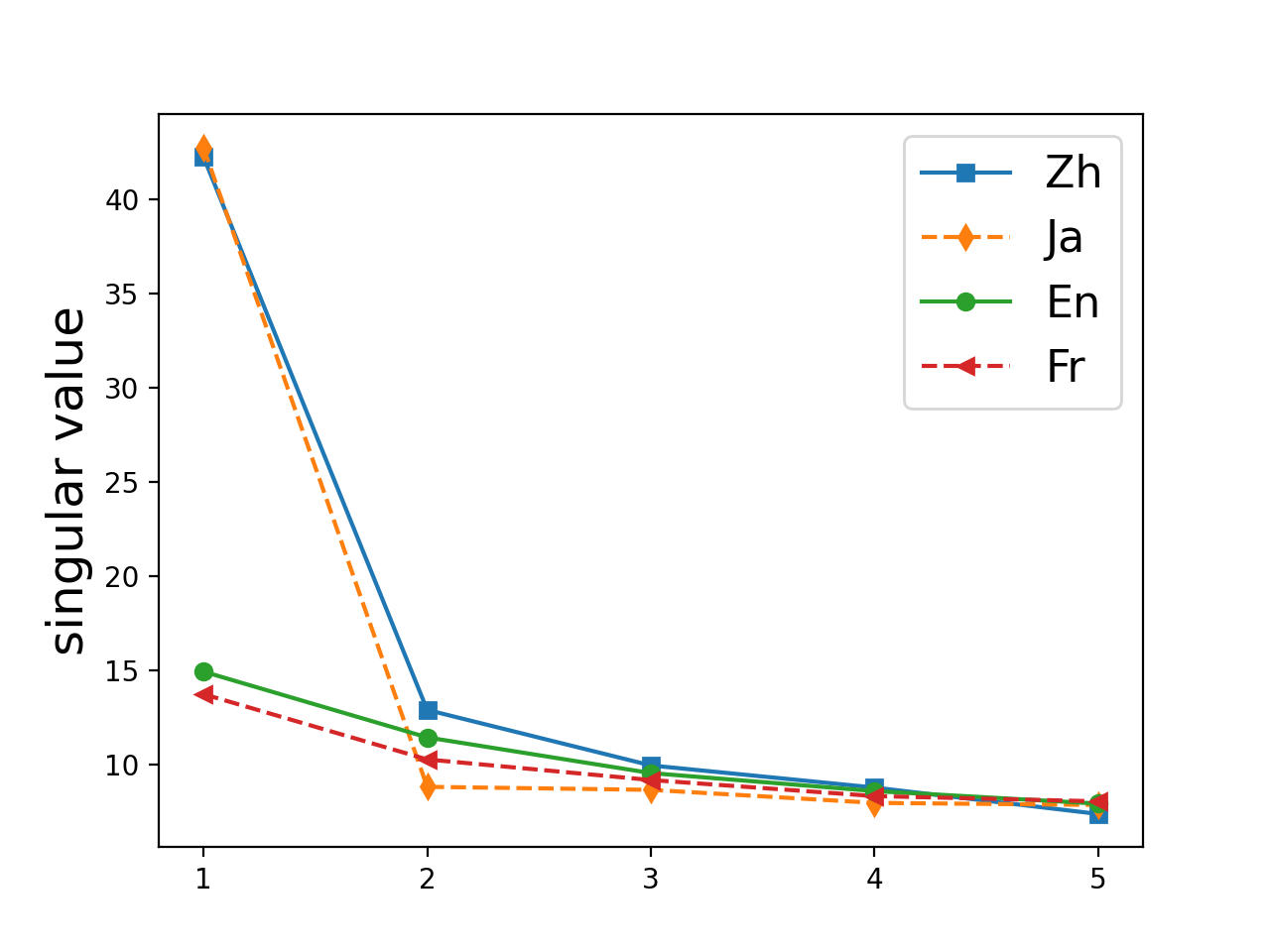}
\vspace{-0.4cm}
\caption{Top-5 singular values of original embedding matrices.}
\label{5svd}

\setlength{\abovecaptionskip}{-5cm}
\end{figure}

Figure~\ref{5svd} shows top-5 singular values of original embedding matrices of various languages. 
The singular values of En and Fr are approximately equal. While the largest singular value of Chinese and Japanese are much larger than that of English. So it is necessary to make them to match with other. 

The phenomena shown in Figure~\ref{5svd} is consistent with the experiments of  \citet{li-etal-2020-simple} who find that the highest eigenvalue of embeddings matrix is much larger than the others in most of the failed languages. Our work is inspired by their observations. The difference is that we bridge the gap by jointly scaling the embeddings and they drop the dimension corresponding to the highest eigenvalue in the initialization step. We are interested in combining the two approaches in the future work to achieve more improvements.
\section{Conclusion}
In this work, we propose a novel method for robust UBLI based on transformation which can make embeddings to match with other.
Our method obtains substantial gains in distant language pairs without scarifying the performance of similar language pairs.
In the future work, we will integrate our method with other models such as MUSE.

\bibliographystyle{acl_natbib}
\bibliography{anthology,acl2021}


\end{document}